\documentclass[letterpaper, 10 pt, conference]{ieeeconf}  

\IEEEoverridecommandlockouts                              

\usepackage[utf8]{inputenc}
\usepackage[T1]{fontenc}
\usepackage{cite}
\usepackage{amsmath,amssymb,amsfonts}
\usepackage{algorithmic}
\usepackage{graphicx}
\usepackage{textcomp}
\usepackage{xcolor}
\usepackage{breqn}
\usepackage{textcase}
\usepackage[export]{adjustbox}
\usepackage{subcaption}
\usepackage{listings}
\usepackage{tabularx}

\title{\LARGE \bf
Parameterisation of lane-change scenarios from real-world data
}
\author{Dhanoop Karunakaran$^{1}$, Julie Stephany Berrio$^{1}$, Stewart Worrall$^{1}$, Eduardo Nebot$^{1}$
\thanks{$^{1}$D.Karunakaran, J.Berrio, S. Worrall,  E. Nebot  are with the Australian Centre for Field Robotics (ACFR) at the University of Sydney (NSW, Australia).
       E-mails: {\tt\small \{d.karunakaran,  j.berrio, s.worrall,  e.nebot\}@acfr.usyd.edu.au}}
}

\begin{document}

\maketitle
\thispagestyle{empty}
\pagestyle{empty}

\begin{abstract}

Recent Autonomous Vehicles (AV) technology includes machine learning and probabilistic techniques that add significant complexity to the traditional verification and validation methods. 
The research community and industry have widely accepted scenario-based testing in the last few years. As it is focused directly on the relevant crucial road situations, it can reduce the effort required in testing. 
Encoding real-world traffic participants' behaviour is essential to efficiently assess the System Under Test (SUT) in scenario-based testing. So, it is necessary to capture the scenario parameters from the real-world data that can model scenarios realistically in simulation. 
The primary emphasis of the paper is to identify the list of meaningful parameters that adequately model real-world lane-change scenarios. 
With these parameters, it is possible to build a parameter space capable of generating a range of challenging scenarios for AV testing efficiently.
We validate our approach using Root Mean Square Error(RMSE) to compare the scenarios generated using the proposed parameters against the real-world trajectory data. In addition to that, we demonstrate that adding a slight disturbance to a few scenario parameters can generate different scenarios and utilise Responsibility-Sensitive Safety (RSS) metric to measure the scenarios' risk.

\end{abstract}

\section{Introduction}

Many traditional validation approaches in the automobile industry, such as test matrix, are insufficient for assessing the advanced functionalities efficiently due to the use of simple and repeated non-challenging scenarios\cite{zhao2016accelerated}. Road testing is a widespread and essential testing approach to validate Advanced Driver System(ADS)\cite{waymo2017road}. However, most driving data collected during road testing is considered non-critical. Thus most of the scenarios are not relevant enough for evaluation purposes\cite{xinxin2020csg}. Scenario-based testing is widely accepted in the research community and automotive industry that complement the road testing approach\cite{de2017assessment, leitner2020enable,putz2017system}. This approach for validating Highly Automated Vehicles (HAVs) is promising as it reduces the test efforts by focusing on meaningful scenarios without the need of driving millions of kilometres \cite{elrofai2016scenario, amersbach2019functional, ponn2020identification}.

In scenario-based testing, generating events for assessing the system is crucial. There are two types of scenario generation: knowledge-driven and data-driven\cite{hauer2020clustering,najm2007pre}. A knowledge-driven approach can generate scenarios simply and cost-effectively using expert knowledge. At the same time, a data-driven model extracts the scenarios from real-world data. Even though the data-driven approach is time-consuming, it encodes in form of parameters the behaviour of the real-world traffic participants that is essential for validating the ADS. 

There are many projects such as PEGASUS and ENABLES3 published as part of the collaboration of the academia and industry\cite{winner2019pegasus,leitner2020enable}. The main component of these projects is the scenario-based safety assessment methodology. It is fundamental to capture scenarios from real-world data that represents realistic traffic participants' behaviour. The scenario extraction parameterises the real-world traffic information. These values are used as a baseline to generate many scenarios in simulation for assessing the System Under Test(SUT). 

United Nations Economic Commission for Europe (UNECE) regulation no. 157\cite{united2020proposal} is one of the first regulatory frameworks for the approval of advanced functionality in automobiles. The regulation is involved in the approval of Automated Lane Keeping Systems(ALKS), and it utilises scenario-based testing for the assessment. The framework defines the set of parameters to define the lane change scenarios such as cut-in, cut-out, and deceleration. However, the regulation lacks the required parameterisation necessary that reduces the ability to capture the behaviour of the traffic participants well. \cite{tenbrock2021conscend} proposed a dataset with concrete scenarios that are derived from real-world data and a list of parameters to model the real-world scenarios. The author argues that UNECE regulation is missing the required parameterisation to represent the real-world scenarios in simulation. They have proposed a set of parameters for the lane change scenarios. In our experiment, it is clear more parameters required to model the real-world scenario better. 

National Highway Traffic Safety Administration (NHTSA) conducted an extensive study on the crash scenarios based on the 2004 General Estimates System (GES) crash database\cite{najm2007pre}. They found that the lane change scenario accounts for 7.62$\%$ of Two-Vehicle Light-Vehicle crashes, equivalent to 295,000 crashes. Given the significance of these maneuvers it is important to generate many and diverse challenging lane change scenarios to evaluate System Under Test(SUT).

It is essential to assess SUT in dangerous scenarios before deploying them on public roads as the collected real-world data may not contain risky scenarios commonly. Extracting the adequate parameters and building the parameter space enables the creation of many critical scenarios by sampling from the parameter space. It allows us to evaluate SUT in more critical scenarios. The absence of suitable parameters reduces the ability to encode the traffic participants' behaviour from real-world data. This paper proposes a list of parameters for cut-in and cut-out scenarios that better model the real-world lane-change scenarios in simulation. It is a crucial step before building the parameter space. We use two metrics, RMSE for longitudinal position and RMSE for lateral position to efficiently check the similarity of generated trajectories with a corresponding real-world scenario. 

We also demonstrate how to generate many variations of the original scenario by adding slight disturbances to a few scenario parameters. The riskiness of the generated scenarios is measured using Intel Mobileye's Responsibility-Sensitive Safety(RSS)\cite{mobileye_rss}. It indicates how dangerous is a particular scenario so that we can use them to evaluate SUT's performance.

\section{Related work}

The United Nations Economic Commission for Europe is a regional commission of the United Nations. It promotes economic cooperation and integration among European countries in particular. They have proposed a regulatory framework for the ALKS approval before deploying on public roads, called UNECE regulation no. 157. It is one of the first regulatory frameworks for assessing automated driving systems. As per the regulation, ALKS can be activated under certain conditions and operational up to the maximum speed of 60km/hr.

The regulation provides a set of parameters to create the test scenarios for assessing the ALKS. In the regulation the scenario starts from the where the lane change maneuver begins. In our case, we start the scenario few seconds before the cut-in or cut-out happens. This is done because the SUT needs initialization before any maneuver starts in the simulation.
Besides, our approach add extra parameters which help to better model the real-world events.

In \cite{tenbrock2021conscend}, the authors have proposed a lane-change scenario extraction strategy to build a dataset and a list of parameters to model real-world scenarios for ALKS as shown in  table~\ref{table:table_cut_in_other_paper}.
They used the HighD dataset \cite{krajewski2018highd} to extract the parameters. Instead of starting the scenario from when the lane change starts as in the regulation \cite{united2020proposal}, they set the start of the scenario 5 seconds before the lateral movement begins. Authors assume the initial constant velocity for all the vehicles, and this value is calculated based on the lane-change maneuver starts in 5 simulation seconds. It does not reflect the behaviour of the traffic participants accurately. Also, having fewer parameters to construct the vehicle's trajectory increases the longitudinal position error. 

\begin{table}[h!]
\centering
\caption{List of parameters for cut-in scenarios\cite{tenbrock2021conscend}.}
\label{table:table_cut_in_other_paper}
\begin{tabular}{||c c||} 
 \hline
 Parameter & Unit\\ [0.5ex] 
 \hline\hline
 Initial ego velocity & m/s \\ 
 \hline
 Initial challenging vehicle velocity  & m/s \\
 \hline
 Initial distance to challenging vehicle  & m \\
 \hline
 Initial challenging vehicle relative lane & -1,+1 \\
 \hline
 Initial challenging vehicle lane offset & m/s \\
 \hline\hline
 Challenging vehicle cut-in trigger distance & m \\
 \hline
 Challenging vehicle cut-in distance & m\\
 \hline
 Final challenging vehicle velocity & m/s\\
 \hline
 Final challenging vehicle lane offset & m/s \\ [1ex] 
 \hline
\end{tabular}
\end{table}

The list of parameters in \cite{tenbrock2021conscend}'s work include velocity and lane position for each of the vehicles, lane offset for the challenging vehicle, triggering distance to trigger the lane-change maneuver, and distance travelled(cut distance) during the maneuver. It is important to note that this approach only considers two control points to describe the challenging vehicle trajectory, the start and endpoint.

Our proposed parameters are extracted from four control points or locations from the real-world scenario as shown in the figure~\ref{fig:points}. It is clear from experiments that more parameters can better model real-world events. The proposed parameters for the cut-out and cut-in scenarios are the same, except for the final lane parameter in the cut-out case.  

\section{Lane change scenario extraction}

\begin{figure}[h]
    \centering
    \begin{subfigure}[b]{0.46\columnwidth}
         \centering
         \includegraphics[width=0.8\columnwidth]{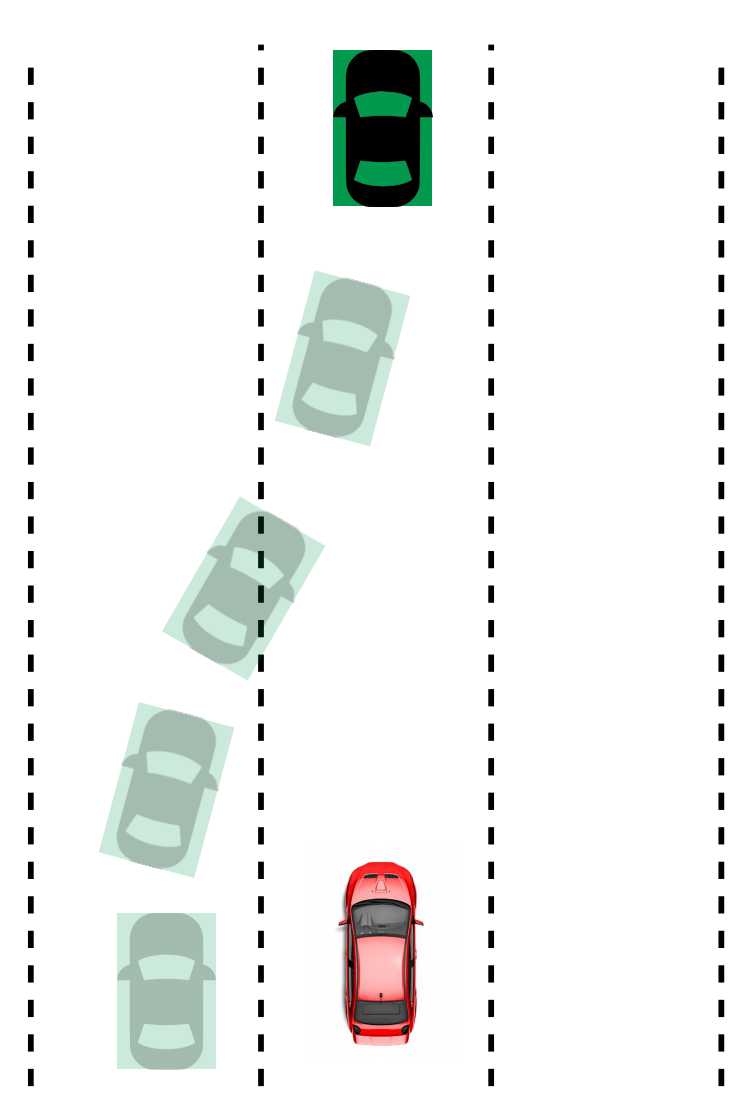}
         \caption{Cut-in scenario: the challenging vehicle changes the lane to ego vehicle's lane. }
         \label{fig:cut_in}
    \end{subfigure}
    \hfill
    \begin{subfigure}[b]{0.46\columnwidth}
         \centering
         \includegraphics[width=0.8\columnwidth]{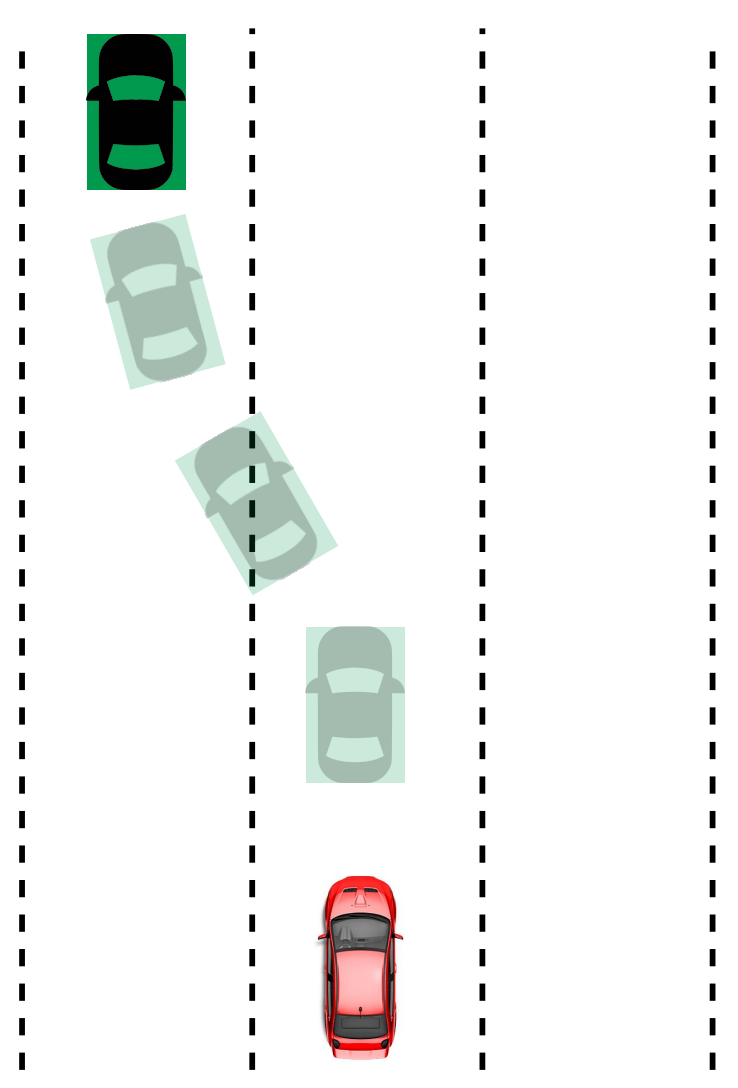}
         \caption{Cut-out scenario: the challenging vehicle changes the lane from ego vehicle's lane. }
         \label{fig:cut_out}
    \end{subfigure}
    \caption{Lane change scenario types, where the ego vehicle is in red, challenging vehicle is in green.}
    \label{fig:lane_change_type}
\end{figure}

In this work, we are interested particularly in two lane-change scenario types, as shown in the figure~\ref{fig:lane_change_type}: cut-in and cut-out. When a challenging vehicle moves into the ego-vehicle lane, it is considered as a cut-in (figure~\ref{fig:cut_in}). Contrary,  cut-out scenario (figure~\ref{fig:cut_out}) happens when the front challenging vehicle changes from the ego-vehicle's to an adjacent lane. 

\begin{figure}[t]
\vspace{2mm}
    \centering
    \includegraphics[width=0.95\columnwidth]{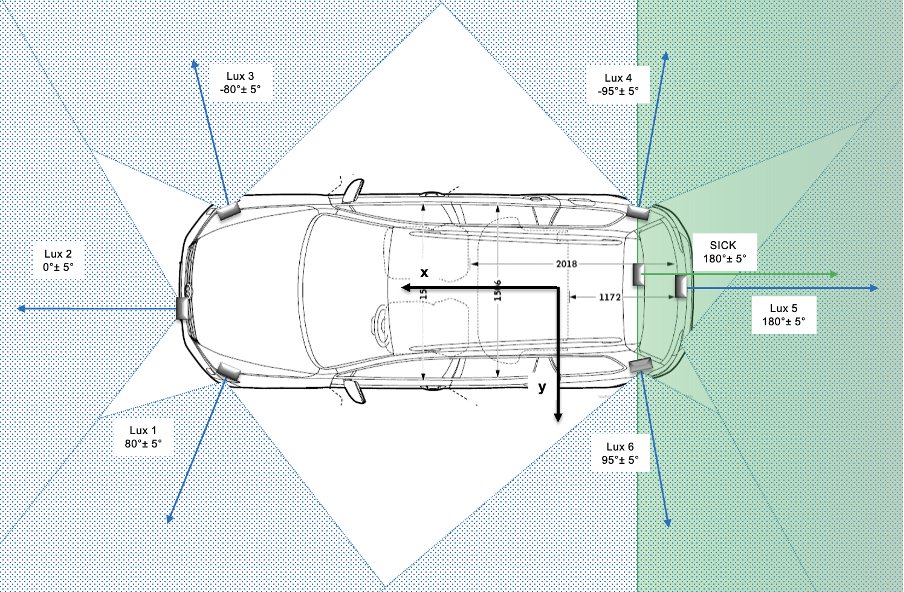}
    \caption{\small Top-down view of the distribution of the data collection vehicle's lidar sensors. In blue is the field of view for the ibeo LUX lidar and in green is the FOV of the SICK lidar. }
    \label{fig:data_collection_vehicle}
\end{figure}

We use the scenario extraction framework from our previous work to capture the parameters proposed in this paper. The data used in this paper to model real-world scenarios was collected using Volkswagen Passat station wagon that is mounted with IBEO and SICK lidars. The placement of the lidars on the vehicle is shown in the Figure~\ref{fig:data_collection_vehicle}. The on-board computer processes the sensor data to detect and track the objects up to 200 meters. 

The first step of the framework constructs the lanes from point cloud data. To segment the road surface from the point cloud, we evaluate height and the first and second derivative of the angle between consecutive points. Then use intensity and range information to cluster close together points with high intensity, which usually belongs to the lane markings. Each cluster is represented by a line segment that connects its extreme points. These line segments are merged if they belong to the same lane. We use the line equations to evaluate the distance between lines at their extreme points, if the distance is less than a threshold, the lines are merged.

\begin{figure}[h]
\vspace{2mm}
\centering
\includegraphics[width=0.8\columnwidth]{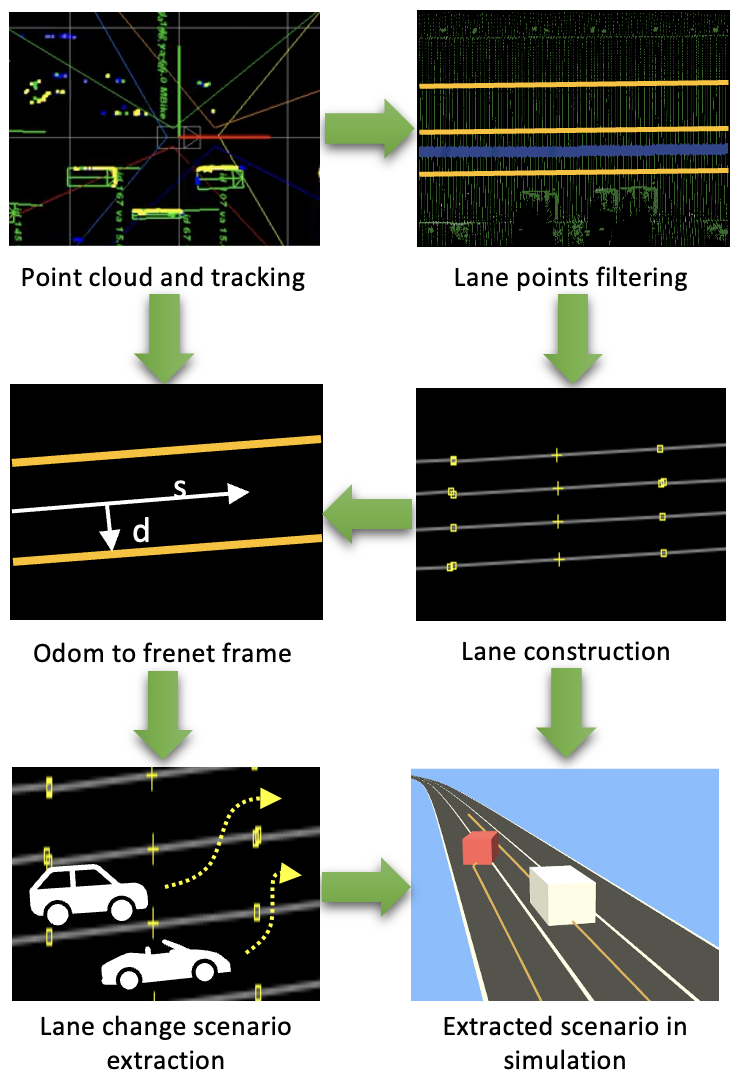}
\caption{\small The figure depicts scenario extraction framework architecture that captures the parameters to model the real-world scenario in simulation. It takes the point cloud, IMU, and object tracking information as input, construct the OpenSCENARIO and OpenDRIVE file using the extracted parameters from the data. We have OpenX files to run in any supported simulation and parameters to build parameter space as the output.}
\label{fig:framework}
\end{figure}

As the second step, we convert all the detected vehicles' positions into to Frenet frame. The Frenet frame has two coordinates: longitudinal displacement $s$, and lateral displacement $t$. We use the ego path as the reference line to compute $s$ coordinate, and then $t$ is the perpendicular distance from the vehicles' position to the reference line. Lane change scenarios are detected by comparing vehicles' lane position and their lateral distance to the ego path at each second. 
Cut-in events are detected as any vehicle in a different lane from to ego vehicle's, which lateral displacement is less than a threshold from the reference path. Likewise, a cut-out scenario is detected when any tracked vehicle changes its lane from the ego vehicle's, that means its lateral distance is greater than a threshold. Figure \ref{fig:framework} depicts the process followed to extract the cut-in cut-out events.

This framework extracts the list of parameters described in the next section from the detected scenarios and create OpenX files. 
The scenario trajectories are represented in OpenSCENARIO \cite{openscenario} format and road network in OpenDRIVE \cite{opendrive} form. The main advantage of using Open Standard formats is the variety of testing tools that can be used. We use OpenSCENARIO player, Esmini to run the generated scenarios and collect data to apply the similarity checks.

\section{Scenario parameters}

The scenario extraction framework extracts the road network information and the list of proposed parameters shown in table~\ref{table:table_cut_in_our}.
It creates the OpenSCENARIO and OpenDRIVE files using extracted parameters and road information. We use an OpenSCENARIO player called Esmini to play the scenarios and capture the data to run a similarity check with real and generated scenarios. In this section, we explain the parameters used to model the real-world lane-changing events.

\subsection{Proposed parameters}

Our proposed parameters for cut-in and cut-out scenarios parameters are the same, except for the challenging vehicle's final lane number in the cut-out scenario.
A lane-change maneuver is initiated for both scenario types when the triggering longitudinal distance is met. In the cut-in scenario, the final lane of the challenging vehicle corresponds to the ego vehicle's, while in cut-out events the challenging vehicle changes its lane to an adjacent one.

\begin{figure}[h]
\vspace{2mm}
\centering
\includegraphics[width=0.95\columnwidth]{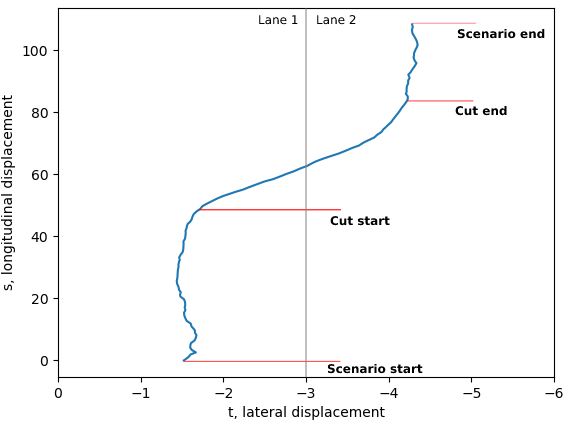}
\caption{\small We extracts parameters in four control points from the real-world data to model the event in simulation.  We call the scenario generated '4 points' scenario.}
\label{fig:points}
\end{figure}

There are two parameters for the ego-vehicle which initialise the vehicle. The \textit{'initial ego velocity'} parameter sets the vehicle's starting velocity. The \textit{'initial ego absolute lane number'} parameter is used to place the ego-vehicle in a specific lane. 
We do not need to add all the parameters used to create the ego-vehicle's trajectory as the proposed parameters as the vehicle will be replaced by SUT. For instance, if we want to test the Emergency Braking System (EBS), we will let run the whole system in one of the generated scenarios to evaluate its efficacy.
 
We extract the parameters of challenging vehicles in four control points from real-world data. These  control points are shown in Figure\ref{fig:points}: scenario\_start, cut\_start, cut\_end, and scenario\_end.

\begin{itemize}
    \item The scenario\_start point includes the initial velocity and the absolute lane number to define the starting lane of the vehicle. The initial distance is a relative and longitudinal distance between ego and challenging vehicles to define their starting position. In OpenSCENARIO, we use these parameters to initialise the challenging vehicle.
    
    \item We extract three parameters each from cut\_start and cut\_end points in the real-world scenario: velocity, total distance, and time travelled from the previous to the current checkpoint. The velocity parameter is used in OpenSCENARIO to interpolate the vehicle's velocity up to that value when the vehicle reaches the distance travelled. We use the time parameter to define the dynamic shape component in OpenSCENARIO to create a velocity profile.
    
    \item Five parameters of the challenging vehicle are captured from the scenario\_end point in the real-world data. Three parameters describe: the final velocity, total distance travelled and duration to reach the end of the scenario from the cut\_end. As in the earlier explanation, these parameters are used to set the trajectory of the challenging vehicle in OpenSCENARIO. The 'final challenging vehicle lane offset' parameter is used to set the lateral position of the challenging vehicle from the lane centre. Only for cut-out scenarios, we set a parameter representing the challenging vehicle's final lane number.
    
\end{itemize}

\begin{table}[h]
\centering
\caption{Proposed meaningful parameters for cut-in and cut-out scenarios.}
\label{table:table_cut_in_our}
\begin{tabular}{||c c||} 
 \hline
 Parameter & Unit\\ [0.5ex] 
 \hline\hline\hline
 Initial ego velocity & m/s \\ 
 \hline
 Initial challenging vehicle velocity  & m/s \\
 \hline
 Initial distance to challenging vehicle  & m \\
 \hline
 Initial ego absolute lane number & -1 to -4 \\
  \hline
 Initial challenging vehicle absolute lane number & -1 to -4 \\
 \hline
 Initial challenging vehicle lane offset & m \\
 \hline
 Challenging vehicle trigger distance & m \\
 \hline
 Challenging vehicle velocity at cut start  & m/s\\
 \hline
 Challenging vehicle distance travelled at cut start & m\\
 \hline
 Challenging vehicle duration from start to cut start & s\\
 \hline
 Challenging vehicle velocity at cut end  & m/s\\
 \hline
 Challenging vehicle distance travelled at cut end & m\\
 \hline
 Challenging vehicle duration from cut start to cut end & s\\
 \hline
 Final challenging vehicle velocity & m/s\\
 \hline
 Total distance travelled & m\\
 \hline
 Challenging vehicle duration from cut end to scenario end & s\\
 \hline
 Challenging vehicle cut distance & m\\
 \hline
 Final challenging vehicle lane offset & m \\
 \hline
 Final challenging vehicle lane number & -1 to -4 \\
 \hline
\end{tabular}
\end{table}

There are two other parameters: trigger distance and cut distance. Both values are used to define the lane change maneuver. The first parameter is the relative distance between ego and challenging vehicle to trigger the lane change action. The second parameter defines how long the challenging vehicle has to travel longitudinally during the lane change maneuver.

During the OpenSCENARIO file generation, we use the extracted parameters from the initial point, such as initial velocity, position, lane offset, and lane number parameters, to define the objects' initial behaviour. We use condition and action to create an event in OpenSCENARIO. For instance, a lane change maneuver such as cut-in and cut-out event for the challenging vehicle is generated using relative distance conditions and lane change action. We use the triggering\_distance parameter for the condition and the final\_lane parameter for the lane change action. During the execution of the OpenSCEANARIO file, the action is executed when the relative and longitudinal distance condition is met. The vehicles follow similar trajectories as in real-world data to do the lane change action to meet the above condition. We have used velocity, distance travelled, and duration parameters to create a similar trajectory. We start with a cut\_start point where we create events with the OpenSCENARIO's 'travelled distance' condition with distance parameter, and the velocity parameter is used for the OpenSCENARIO's 'absolute speed' action. The duration parameter fills the transition dynamics of the 'absolute speed' action. We repeat the event creation for two other checkpoints: cut\_end and scenario\_end. When we run the OpenSCENARIO file, these events are triggered to apply then actions when they meet the travelled distance condition.

\cite{tenbrock2021conscend} considers parameters from two control points instead of four points: scenario\_start and scenario\_end to model the trajectory. They have the same two parameters to represent the lane change maneuver: triggering distance and cut distance. The list of parameters are listed in the table ~\ref{table:table_cut_in_other_paper}. 
We use two terminologies to differentiate the scenarios generated from different parameters in the plot. Firstly, the 'four points' scenarios represent the generated scenarios using our proposed parameters, and the 'two points' scenario is the terminology for the generated scenarios using parameters in the other work\cite{tenbrock2021conscend}.
Table \ref{table:table_cut_in_our} summarises the proposed parameters for cut-in and cut-out scenarios.

\subsection{Metrics}

\begin{figure*}[t]
\vspace{2mm}
    \centering
    \hfill
    \begin{subfigure}[b]{0.44\columnwidth}
         \centering
         \includegraphics[width=0.99\columnwidth]{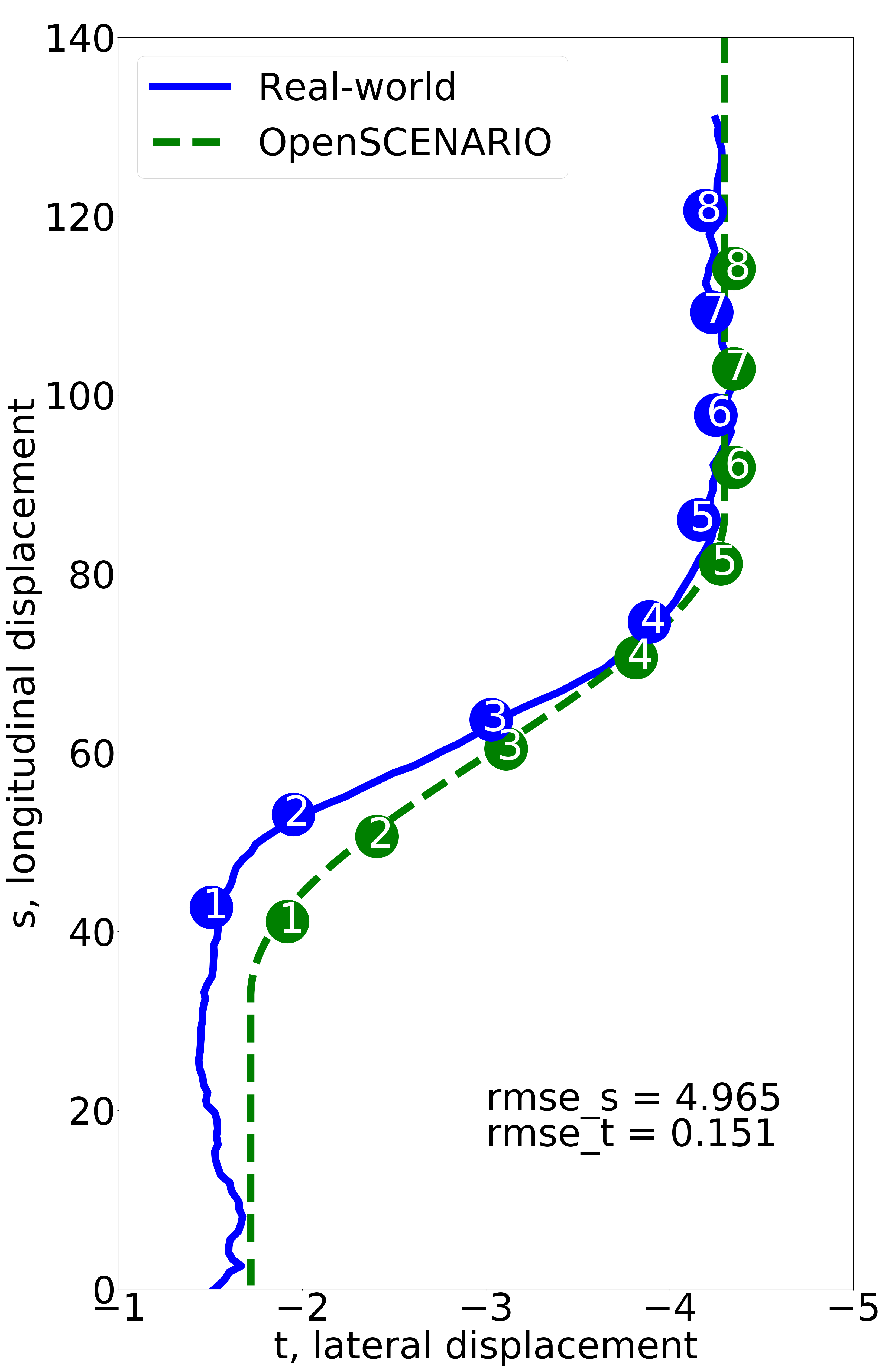}
         \caption{2 points first scenario}
         \label{fig:scenario_2_points_1}
    \end{subfigure}
    \hfill
    \begin{subfigure}[b]{0.44\columnwidth}
         \centering
         \includegraphics[width=0.99\columnwidth]{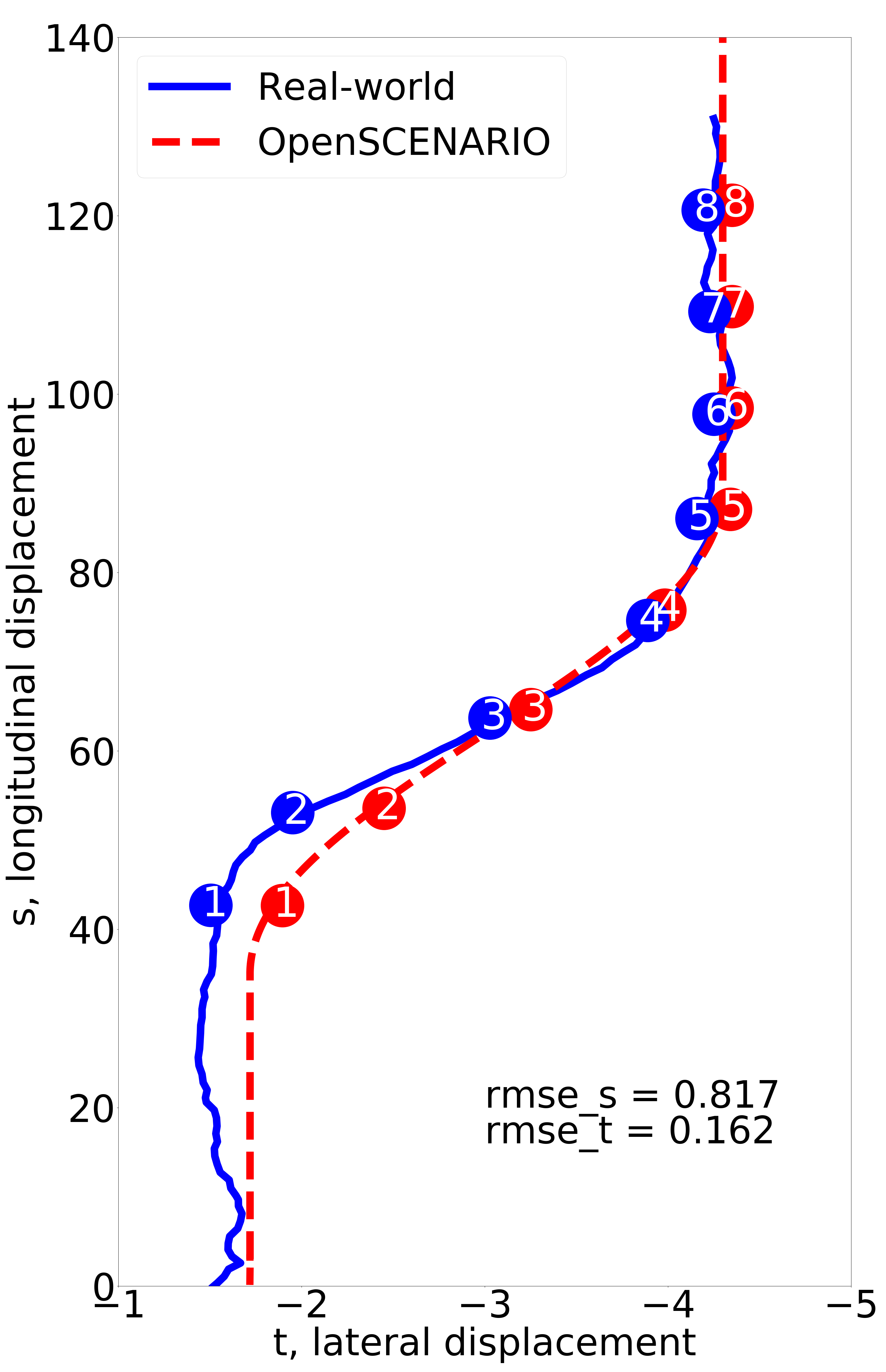}
         \caption{4 points first scenario}
         \label{fig:scenario_4_points_1}
    \end{subfigure}
    \hfill
    \begin{subfigure}[b]{0.44\columnwidth}
         \centering
         \includegraphics[width=0.99\columnwidth]{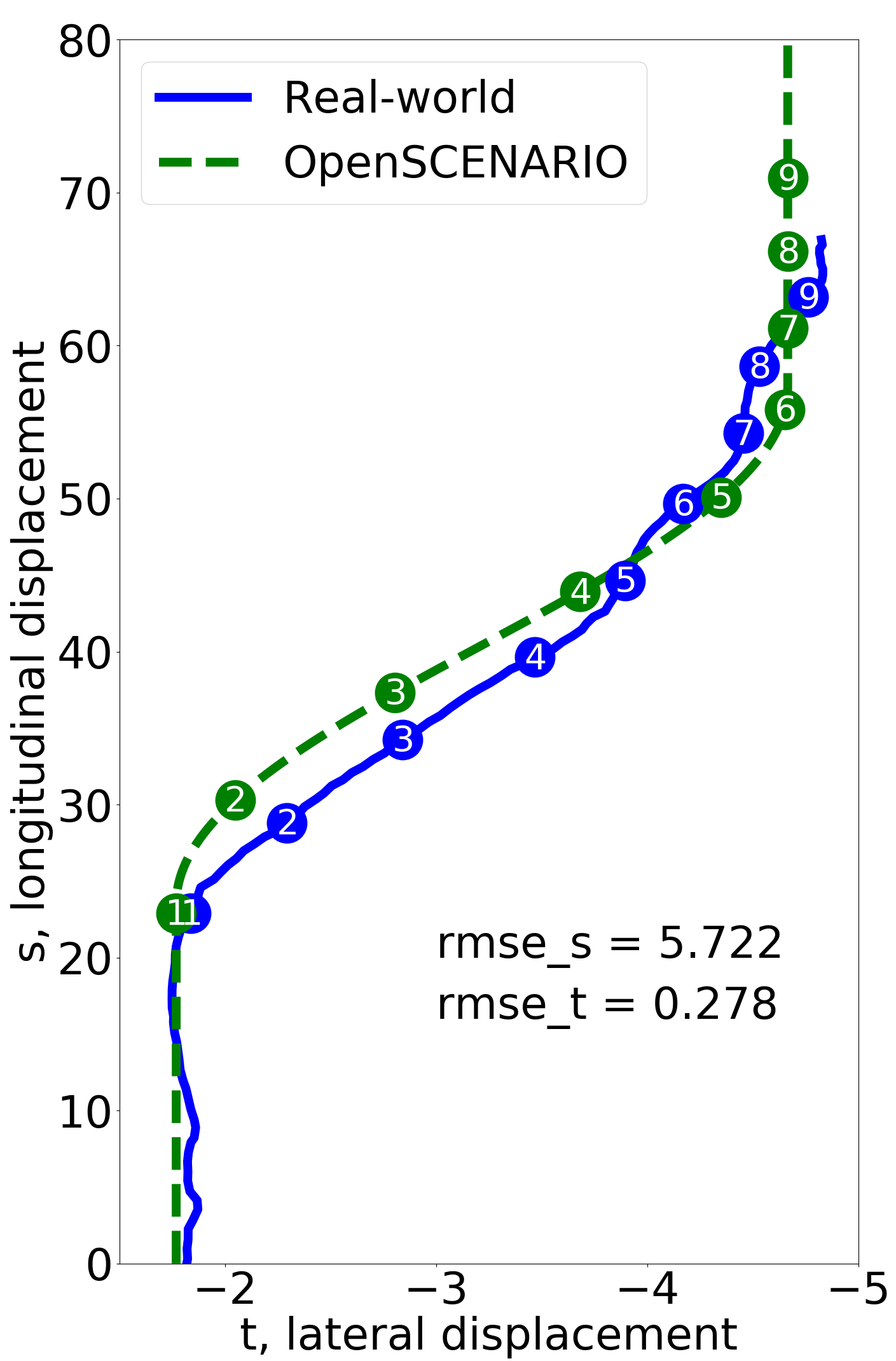}
         \caption{2 points second scenario}
         \label{fig:scenario_2_points_2}
    \end{subfigure}
    \hfill
    \begin{subfigure}[b]{0.44\columnwidth}
         \centering
         \includegraphics[width=0.99\columnwidth]{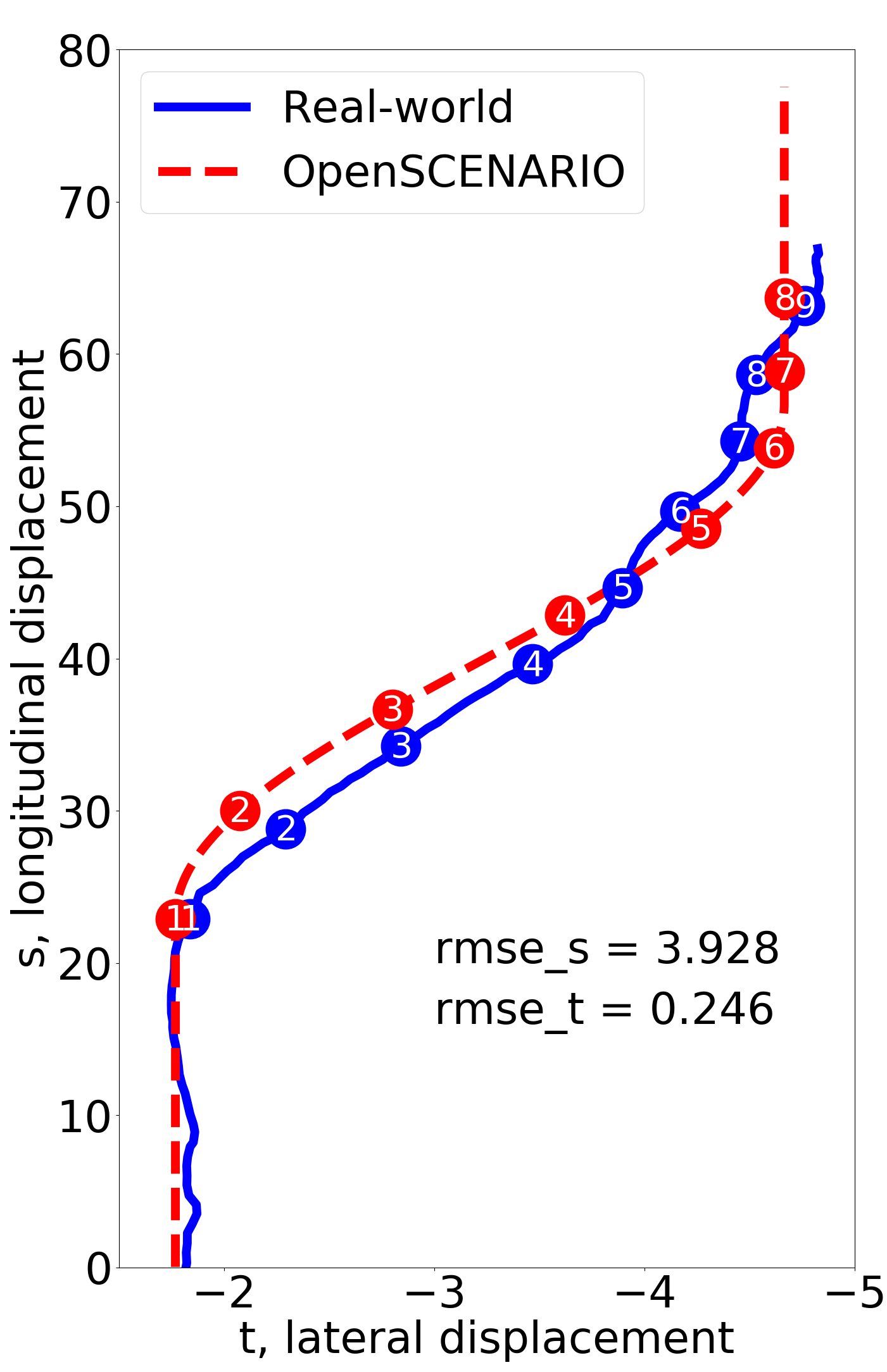}
         \caption{4 points second scenario}
         \label{fig:scenario_4_points_2}
    \end{subfigure}
    \caption{
Plots show the temporal (seconds from the cut start within circles) and spatial alignment of generated scenarios of two real-world scenarios. It confirms that the temporal aspect 
of the '4 points' scenario is well-aligned with the real-world scenario. Also, the RMSE indicates the less longitudinal and lateral error value for the '4 points' scenario than the '2 points' scenario. Even though the '2 points' scenario looks similar, the alignment in time for our 4 point model is considerably more accurate.}
    \label{fig:scenario1}
\end{figure*}

The scenario extraction framework captures the proposed parameters and generates OpenSCENARIO and OpenDRIVE files. In order to check the similarity, we run the generated OpenX files in the Esmin, OpenSCENAIRO player, to collect the generated scenario data. We use two metrics to identify the similarity of real-world trajectory with generated trajectory: Root Mean Square Error(RMSE) for longitudinal position and Root Mean Square Error(RMSE) for the lateral position. 

As shown in the equation~\ref{eq1}, we take the difference between real longitudinal position, $d_s^{real}$ that is taken from real-world data and longitudinal position, $d_t^{gen}$ from the generated scenario at each time step. The timestep in this experiment is one second. Then we take the squared value of the difference to get the absolute value. The mean of squared error is computed by dividing the sum of squared errors by total timesteps. Finally, taking the square root gives the RMSE for the longitudinal position.
\begin{equation}
\label{eq1}
rmse_{s|t}= \sqrt{  \frac{\sum_{0}^{n-1}(d_{(s|t)n}^{real}-d_{(s|t)n}^{gen})^2}{n}}
\end{equation}
RMSE for the lateral position is the same as the longitudinal position, except we replace the longitudinal position, $s$, with lateral position $t$. We compute both RMSE from the start of the lane-change maneuver until the scenario's end.

In the literature RMSE for lateral position proposed for cut-in scenario\cite{tenbrock2021conscend}. However, this metric is not enough to check the similarity. The longitudinal error is higher if the trajectory generated is not similar to the real scenario, even though the lateral position error is low. 


We make use of Intel Mobileye's Responsibility-Sensitive Safety(RSS)\cite{mobileye_rss} to measure how dangerous is the generated scenarios. The ultimate goal of RSS as a formal method is to guarantee that an agent will not cause an accident, rather than to guarantee that an agent will not be involved in an accident. A safe distance is calculated longitudinally and laterally based on the formulas provided by the method. This method considers the worst-case scenario, eliminating the need to estimate road user intentions. This paper focuses on the longitudinal RSS that gives the minimum safe distance required longitudinally. The longitudinal RSS is calculated using the speed of ego-vehicle and lead vehicle with other constants in the RSS equation~\ref{rss_eq1}. 

\begin{equation} \label{rss_eq1}
d_{min}=\left[v_r\rho+\frac{1}{2}a_{max,a}\rho^2+\frac{(v_r+\rho a_{max,a})^2}{2a_{min,b}}-\frac{v_f^2}{2a_{max,b}}\right]
\end{equation}

where $\ d_{min} $ represents the longitudinal safe distance, $\ v_r $  and $\ v_f $ are the velocity of the agent vehicle and front vehicle, respectively. $\ a_{min,b}$ is the minimum reasonable braking force of the agent vehicle, $\ a_{max ,b}$ is the maximum braking force of the front vehicle. In terms of acceleration, $\ a_{max,a} $ is the maximum acceleration of the front vehicle. $\rho$ is the agent vehicle response time. We use default values for these constants from the c++ library for the RSS\cite{rss_c++}.

\section{Results}


The extraction framework captures the lane-change scenario parameters listed in the table~\ref{table:table_cut_in_our} from real-world data. The framework uses point cloud, IMU, and object tracking data to capture the parameters and road network information. Then it generates OpenSCENARIO and OpenDRIVE files to regenerate the real-world scenarios in simulation. We use Esmini, an OpenSCENARIO player, to capture the data in simulation to run a similarity check with real-world data. We run the similarity checks using RMSE between generated and real-world scenarios. 
In addition, we demonstrate the generation of different scenarios by adding disturbance to a few parameters and measure how risky is the newly generated scenario is using RSS. As described in the parameter section, we have two generated cut-in scenarios: a '4 points' scenario and a '2 points' scenario. In the plots, we have added circles showing the second starts from the cut-in maneuver in which the vehicle reaches that position.


\begin{figure*}
\vspace{2mm}
\def\tabularxcolumn#1{m{#1}}
\begin{tabularx}{\linewidth}{@{}cXX@{}}
\begin{tabular}{cc}
\subfloat[]{\includegraphics[width=4.6cm, height=4.1cm]{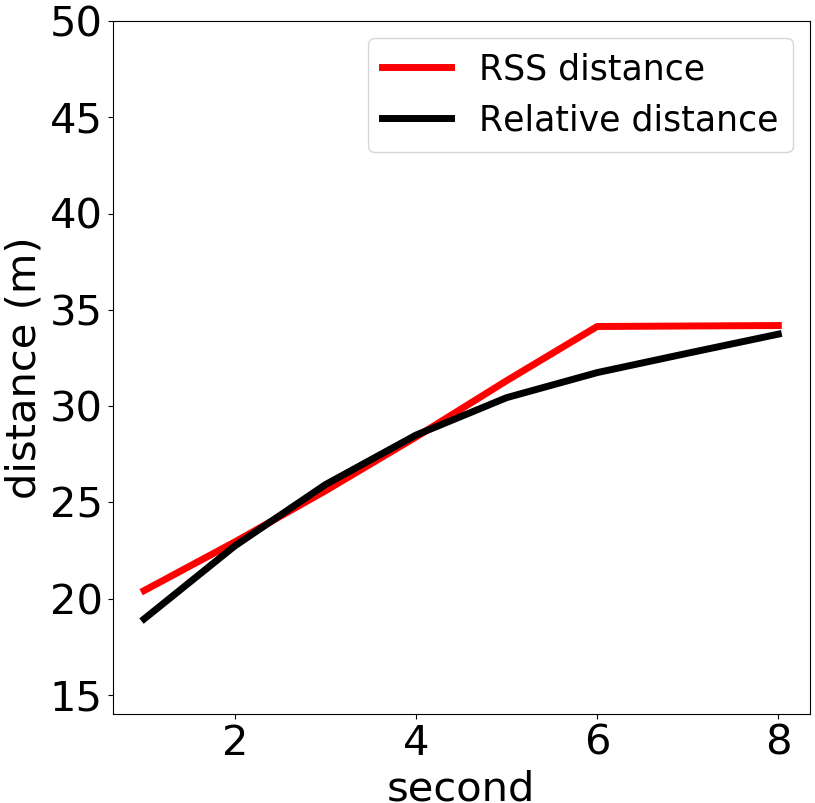}\label{fig:rss_without_noise}} 
   & \subfloat[]{\includegraphics[width=4.6cm,height=4.1cm]{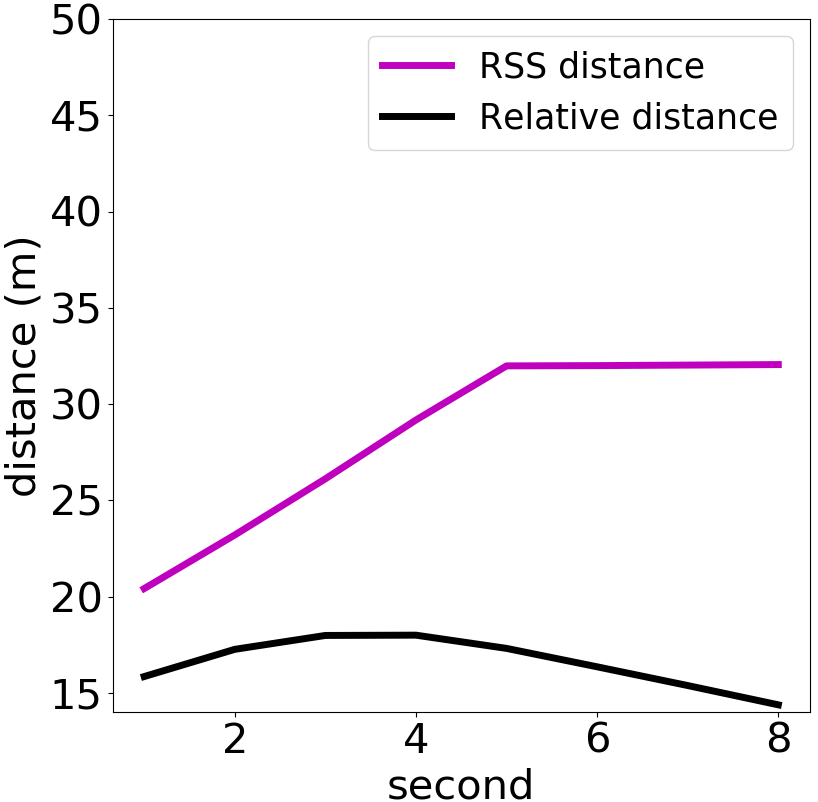}\label{fig:rss_noise1}}\\
\subfloat[]{\includegraphics[width=4.6cm,height=4.1cm]{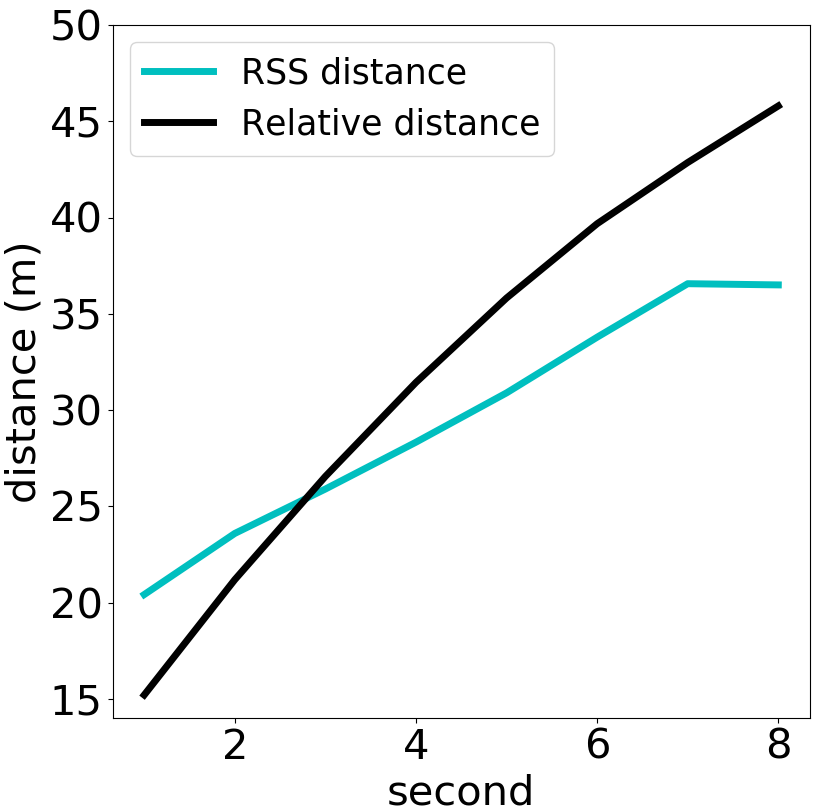}\label{fig:rss_noise2}} 
   & \subfloat[]{\includegraphics[width=4.6cm,height=4.1cm]{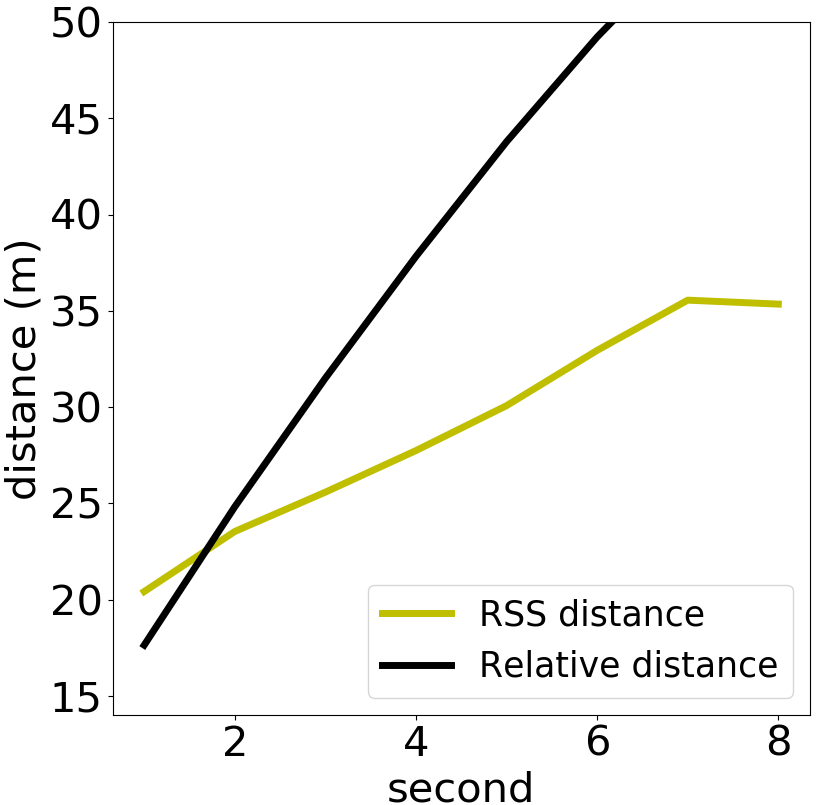}\label{fig:rss_noise3}}\\
\end{tabular}
&
\subfloat[]{\includegraphics[width=6.6cm,height=8.7cm]{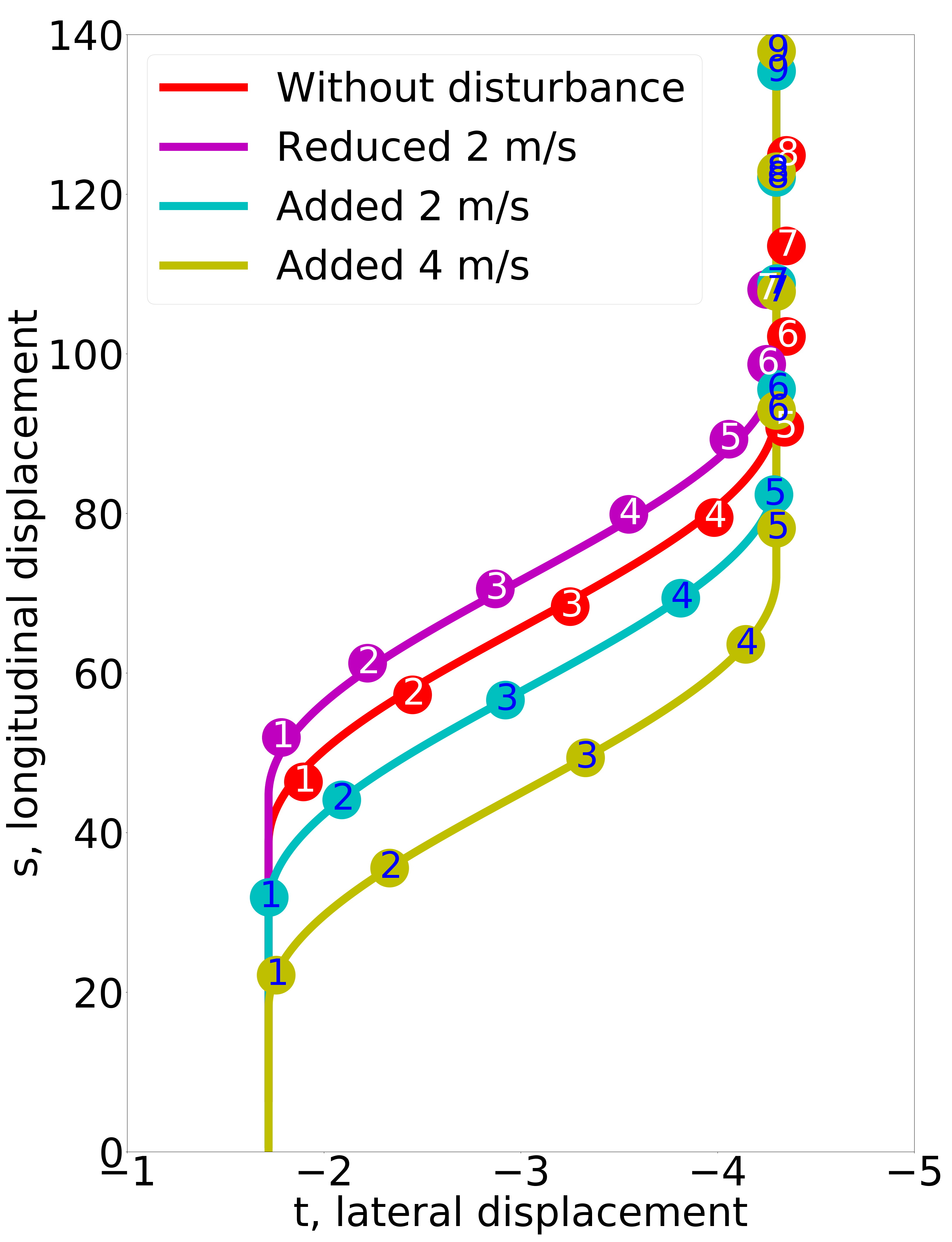}\label{fig:s_t_noice}}
\end{tabularx}

\caption{This figure shows how we can generate new trajectories with different risk properties by adding disturbances to the '4 points' model. (i.e. (a) is the original scenario, (b) is risky scenario, (c) and (d) are more conservative scenarios. Figure e shows the trajectory of various generated scenarios by adding disturbances.}\label{fig:scenario_noice}
\end{figure*}

The scenarios plotted in the figure~\ref{fig:scenario1} shows the comparison of generated and real-world scenarios. The figure was constructed with two real-world scenarios to compare.The figure~\ref{fig:scenario_2_points_1} and figure~\ref{fig:scenario_4_points_1}  shows the comparison of first scenario. In the figure~\ref{fig:scenario_4_points_1}, our proposed parameters are used to model the event that creates well aligned in time and distance travelled longitudinally to the real-world scenario. The generated scenario has a minimal RMSE  of 0.817 longitudinally and 0.162 laterally. Minimal longitudinal RMSE is evident in the plot~\ref{fig:scenario_4_points_1} that the longitudinal position travelled at each second from the start of the cut-in maneuver is similar. Whereas 'two points' scenario generated using the same real-world data as shown in the figure~\ref{fig:scenario_2_points_1} has RMSE of 4.95 longitudinally and a 0.15 laterally. The '2 points' model looks the same in the plot, but the '4 points' scenario model is better aligned in time and distance. It indicates that the '4 points' approach can reproduce the real-world scenario in simulation better than the '2 points' scenario model. We can see a similar result in the second real-world scenario and generated scenarios plotted in the figure~\ref{fig:scenario_2_points_2} and figure~\ref{fig:scenario_4_points_2}. The 'four points' scenario has recorded 3.92 longitudinal RMSE and lateral RMSE of 0.246.
In contrast, the '2 points' scenario has a higher longitudinal RMSE of 5.72. Compared to the previous example shown in the figure~\ref{fig:scenario_4_points_1}, the '4 points' model has higher longitudinal RMSE. It is due to the longer time to complete the lane-change maneuver, leading to the challenging vehicle's velocity profile not matching exactly to the real-world one. Even though the '4 points' model has the highest longitudinal RMSE, the model can represent the real-world scenario better than the '2 points' scenario.

The figure~\ref{fig:scenario_noice} depicts a generated cut-in scenario by adding disturbances to a few parameters from the initially generated scenario(figure~\ref{fig:scenario_4_points_1}).
This experiment demonstrates how we can generate scenarios with different risk properties by varying a few parameters. We changed values from three velocities parameters:'Challenging vehicle velocity at the cut start', 'Challenging vehicle velocity at the cut end', and 'Challenging vehicle final velocity'. The figure~\ref{fig:s_t_noice}
shows the different scenarios generated with various disturbances. 
The red coloured line in the figure represents the '4 points' model. We add disturbance to this model to generate three other scenarios. When we reduce two m/s from the three velocity parameters to the '4 points' scenario model, the relative velocity between the ego and the challenging vehicle gets reduced. As the triggering distance is a relative longitudinal distance between ego and challenging vehicle, reducing relative velocity means travelling longer to meet the triggering distance. This behaviour is plotted in the 'magenta' coloured line, as shown in the figure~\ref{fig:s_t_noice}. However, adding two m/s resulted in a different scenario, as shown in the 'cyan' coloured line. It is expected behaviour
as adding more velocity to the challenging vehicle parameters leads to the higher relative velocity between ego and challenging vehicle. It triggers the lane change maneuver earlier than the '4 points' model because it meets the triggering distance faster. Similarly, we add four m/s to the three parameters, resulting in a much quicker lane-change maneuver. Changing parameters in the scenario can generate variations of the original scenario to evaluate SUT. We showed that it could be done with just three parameters through this experiment.


We also measure how dangerous is the newly generated scenario using longitudinal RSS and the relative distance between ego and challenging vehicles. We compute longitudinal RSS and relative distance from the lane change maneuver to the end of the scenario. As the RSS distance gives the minimum safe distance required to keep at any given time, the scenario is considered risky when the relative distance is lower than the RSS distance for a long time. 
Figure~\ref{fig:rss_without_noise} depicts the comparison of RSS and relative distances of the '4 points' model. Most of the time, the relative distance is closer to the minimum safe distance computed and towards the end, the total distance is slightly lesser than the RSS distance. The scenario is not dangerous as the relative distance is closer to RSS. Figure~\ref {fig:rss_noise1} shows the RSS and relative distance of the scenario generated by reducing two m/s from the velocity parameters. The figure illustrates that the relative distance is lower than the minimum safe distance required to keep by RSS calculation. With a lower relative distance than the minimum safe distance to keep, an accident can occur at any time due to the sudden acceleration of ego-vehicle and deceleration of the challenging vehicle(lead vehicle). It is a risky scenario generated from a relatively safe scenario by adding a slight disturbance to three velocity parameters
As shown in the 'cyan ' and 'yellow' coloured lines in the figure~\ref{fig:s_t_noice}, adding the velocity to the generated scenario from real-world data can enable a quick lane change. It means that the relative velocity between the ego and the challenging vehicle is higher, resulting in a higher relative distance than the RSS. The figure~\ref{fig:rss_noise2} and figure~\ref{fig:rss_noise3} depicts this behaviour. As in both examples, the challenging vehicle has a higher relative distance than the minimum safe distance required by RSS except in the first two seconds. It means that the longitudinal distance(relative distance) between ego and the challenging vehicle is higher than the minimum safe distance required to keep by the RSS most of the time. The scenarios are non-risky as the quick acceleration of ego-vehicle and deceleration cannot cause any accidents. It is because RSS takes account of acceleration and deceleration to compute the minimum safe distance.



\section{Conclusion and discussion}
The result clearly shows that the list of proposed parameters is more appropriate to model the real-world scenario than fewer parameters. It uses a novel scenario extraction framework to extract the parameters from the point cloud, IMU and tracking data. Using the extraction strategy and list of appropriate parameters,  we can build adequate parameter space for lane change scenarios to generate concrete scenarios to evaluate SUT. Furthermore, we showed that adding a disturbance to a few parameters can vary the generated scenario from real-world data and create critical scenarios. We used RSS as a metric to evaluate how risky the generated scenarios are and compare one of the generated scenarios with new variant scenarios generated by adding disturbances to the parameters. The parameter space required an adequate representation of the real-world scenarios. The paper can initiate the many research works in the right direction to achieve this.

\section*{ACKNOWLEDGMENT}
This  work  has  been  funded  by  the  Australian  Centre  for Field Robotics (ACFR), Insurance Australia Group (IAG), and iMOVE and supported by the  Cooperative  Research  Centres  program,  an  Australian Government initiative.
\bibliographystyle{IEEEtran}
\bibliography{main.bib}

\begin{thebibliography}{10}
\providecommand{\url}[1]{#1}
\csname url@samestyle\endcsname
\providecommand{\newblock}{\relax}
\providecommand{\bibinfo}[2]{#2}
\providecommand{\BIBentrySTDinterwordspacing}{\spaceskip=0pt\relax}
\providecommand{\BIBentryALTinterwordstretchfactor}{4}
\providecommand{\BIBentryALTinterwordspacing}{\spaceskip=\fontdimen2\font plus
\BIBentryALTinterwordstretchfactor\fontdimen3\font minus
  \fontdimen4\font\relax}
\providecommand{\BIBforeignlanguage}[2]{{%
\expandafter\ifx\csname l@#1\endcsname\relax
\typeout{** WARNING: IEEEtran.bst: No hyphenation pattern has been}%
\typeout{** loaded for the language `#1'. Using the pattern for}%
\typeout{** the default language instead.}%
\else
\language=\csname l@#1\endcsname
\fi
#2}}
\providecommand{\BIBdecl}{\relax}
\BIBdecl

\bibitem{zhao2016accelerated}
D.~Zhao, H.~Lam, H.~Peng, S.~Bao, D.~J. LeBlanc, K.~Nobukawa, and C.~S. Pan,
  ``Accelerated evaluation of automated vehicles safety in lane-change
  scenarios based on importance sampling techniques,'' \emph{IEEE transactions
  on intelligent transportation systems}, vol.~18, no.~3, pp. 595--607, 2016.

\bibitem{waymo2017road}
L.~Waymo, ``On the road to fully self-driving,'' \emph{Waymo Safety Report},
  pp. 1--43, 2017.

\bibitem{xinxin2020csg}
Z.~Xinxin, L.~Fei, and W.~Xiangbin, ``Csg: Critical scenario generation from
  real traffic accidents,'' in \emph{2020 IEEE Intelligent Vehicles Symposium
  (IV)}.\hskip 1em plus 0.5em minus 0.4em\relax IEEE, pp. 1330--1336.

\bibitem{de2017assessment}
E.~de~Gelder and J.-P. Paardekooper, ``Assessment of automated driving systems
  using real-life scenarios,'' in \emph{2017 IEEE Intelligent Vehicles
  Symposium (IV)}.\hskip 1em plus 0.5em minus 0.4em\relax IEEE, 2017, pp.
  589--594.

\bibitem{leitner2020enable}
A.~Leitner, ``Enable-s3: Project introduction,'' in \emph{Validation and
  Verification of Automated Systems}.\hskip 1em plus 0.5em minus 0.4em\relax
  Springer, 2020, pp. 13--23.

\bibitem{putz2017system}
A.~P{\"u}tz, A.~Zlocki, J.~Bock, and L.~Eckstein, ``System validation of highly
  automated vehicles with a database of relevant traffic scenarios,''
  \emph{situations}, vol.~1, pp. 19--22, 2017.

\bibitem{elrofai2016scenario}
H.~Elrofai, D.~Worm, and O.~O. den Camp, ``Scenario identification for
  validation of automated driving functions,'' in \emph{Advanced Microsystems
  for Automotive Applications 2016}.\hskip 1em plus 0.5em minus 0.4em\relax
  Springer, 2016, pp. 153--163.

\bibitem{amersbach2019functional}
C.~Amersbach and H.~Winner, ``Functional decomposition—a contribution to
  overcome the parameter space explosion during validation of highly automated
  driving,'' \emph{Traffic injury prevention}, vol.~20, no. sup1, pp. S52--S57,
  2019.

\bibitem{ponn2020identification}
T.~Ponn, M.~Breitfu{\ss}, X.~Yu, and F.~Diermeyer, ``Identification of
  challenging highway-scenarios for the safety validation of automated vehicles
  based on real driving data,'' in \emph{2020 Fifteenth International
  Conference on Ecological Vehicles and Renewable Energies (EVER)}.\hskip 1em
  plus 0.5em minus 0.4em\relax IEEE, 2020, pp. 1--10.

\bibitem{hauer2020clustering}
F.~Hauer, I.~Gerostathopoulos, T.~Schmidt, and A.~Pretschner, ``Clustering
  traffic scenarios using mental models as little as possible,'' in \emph{2020
  IEEE Intelligent Vehicles Symposium (IV)}.\hskip 1em plus 0.5em minus
  0.4em\relax IEEE, 2020, pp. 1007--1012.

\bibitem{najm2007pre}
W.~G. Najm, J.~D. Smith, M.~Yanagisawa \emph{et~al.}, ``Pre-crash scenario
  typology for crash avoidance research,'' United States. National Highway
  Traffic Safety Administration, Tech. Rep., 2007.

\bibitem{winner2019pegasus}
H.~Winner, K.~Lemmer, T.~Form, and J.~Mazzega, ``Pegasus—first steps for the
  safe introduction of automated driving,'' in \emph{Road Vehicle Automation
  5}.\hskip 1em plus 0.5em minus 0.4em\relax Springer, 2019, pp. 185--195.

\bibitem{united2020proposal}
U.~N. E.~C. for Europe~(UNECE), ``Proposal for a new un regulation on uniform
  provisions concerning the approval of vehicles with regards to automated lane
  keeping system,'' 2020.

\bibitem{tenbrock2021conscend}
A.~Tenbrock, A.~K{\"o}nig, T.~Keutgens, and H.~Weber, ``The conscend dataset:
  Concrete scenarios from the highd dataset according to alks regulation unece
  r157 in openx,'' in \emph{2021 IEEE Intelligent Vehicles Symposium Workshops
  (IV Workshops)}.\hskip 1em plus 0.5em minus 0.4em\relax IEEE, 2021, pp.
  174--181.

\bibitem{mobileye_rss}
\BIBentryALTinterwordspacing
S.~Shalev{-}Shwartz, S.~Shammah, and A.~Shashua, ``On a formal model of safe
  and scalable self-driving cars,'' \emph{CoRR}, vol. abs/1708.06374, 2017.
  [Online]. Available: \url{http://arxiv.org/abs/1708.06374}
\BIBentrySTDinterwordspacing

\bibitem{krajewski2018highd}
R.~Krajewski, J.~Bock, L.~Kloeker, and L.~Eckstein, ``The highd dataset: A
  drone dataset of naturalistic vehicle trajectories on german highways for
  validation of highly automated driving systems,'' in \emph{2018 21st
  International Conference on Intelligent Transportation Systems (ITSC)}.\hskip
  1em plus 0.5em minus 0.4em\relax IEEE, 2018, pp. 2118--2125.

\bibitem{openscenario}
\BIBentryALTinterwordspacing
{Association for Standardization of Automation and Measuring Systems},
  ``{OpenSCENARIO},'' 2021. [Online]. Available:
  \url{https://www.asam.net/standards/detail/openscenario/}
\BIBentrySTDinterwordspacing

\bibitem{opendrive}
\BIBentryALTinterwordspacing
------, ``{OpenDRIVE},'' 2021. [Online]. Available:
  \url{https://www.asam.net/standards/detail/opendrive/}
\BIBentrySTDinterwordspacing

\bibitem{rss_c++}
B.~Gassmann, F.~Oboril, C.~Buerkle, S.~Liu, S.~Yan, M.~S. Elli, I.~Alvarez,
  N.~Aerrabotu, S.~Jaber, P.~van Beek \emph{et~al.}, ``Towards standardization
  of av safety: C++ library for responsibility sensitive safety,'' in
  \emph{2019 IEEE Intelligent Vehicles Symposium (IV)}.\hskip 1em plus 0.5em
  minus 0.4em\relax IEEE, 2019, pp. 2265--2271.

\end{thebibliography}

\end{document}